\documentclass[conference]{IEEEtran}
\IEEEoverridecommandlockouts
\usepackage{cite}
\usepackage{amsmath,amssymb,amsfonts}
\usepackage{algorithmic}
\usepackage{graphicx}
\usepackage{textcomp}
\usepackage{booktabs}
\usepackage{xcolor}
\def\BibTeX{{\rm B\kern-.05em{\sc i\kern-.025em b}\kern-.08em
    T\kern-.1667em\lower.7ex\hbox{E}\kern-.125emX}}

\IEEEoverridecommandlockouts     
\begin{document}

\title{Local Label-Informed Feature Transfer for\\
       Generating Ground-Truth Medical Images:\\
       A Comparison of GAN- and Diffusion-Based Approaches
}

% \author{\IEEEauthorblockN{Rick Wilming}
% \IEEEauthorblockA{\textit{Physikalisch-Technische Bundesanstalt Berlin}, Germany \\
% rick.wilming@tu-berlin.de}
% \and
% \IEEEauthorblockN{Irem Ozseker, Luca Matteo Cornils}
% \IEEEauthorblockA{\textit{Technische Universität Berlin}, Germany}
% \and
% \IEEEauthorblockN{Ahc\`{e}ne Boubekki, Benedict Clark, Danny Panknin}
% \IEEEauthorblockA{\textit{Physikalisch-Technische Bundesanstalt Berlin}, Germany}
% \and
% \IEEEauthorblockN{Stefan Haufe}
% \IEEEauthorblockA{\textit{Technische Universität Berlin}, Germany \\
% \textit{Physikalisch-Technische Bundesanstalt Berlin}, Germany}
% }

\author{\IEEEauthorblockN{Rick Wilming$^{\ast}$, Irem Ozseker$^{\dagger}$, Luca Matteo Cornils$^{\dagger}$, Ahc\`{e}ne Boubekki$^{\ast}$, \\ Benedict Clark$^{\ast}$, Danny Panknin$^{\ast}$, Stefan Haufe$^{\ast}$$^{\dagger}$ } 
\IEEEauthorblockA{$^{\ast}$\textit{Physikalisch-Technische Bundesanstalt Berlin}, Germany}
\IEEEauthorblockA{$^{\dagger}$\textit{Technische Universität Berlin}, Germany}
\IEEEauthorblockA{rick.wilming@tu-berlin.de}
}

\maketitle

\thispagestyle{plain}
\pagestyle{plain}

\begin{abstract}
Validating Explainable Artificial Intelligence (XAI) methods in medical imaging requires ground-truth data with known locations of informative features.
However, current approaches rely on expert annotations, which are prone to labeling errors, or on hand-crafted artificial perturbations superimposed onto healthy images to mimic lesions or malignant features, which lack clinical realism.
We present Local Label-Informed Feature Transfer (LLIFT), a framework for generating semi-synthetic brain magnetic resonance images with realistic lesions placed in user-controlled regions, which does not require pixel-level lesion annotations during training.
We implement LLIFT with two generative paradigms: LLIFT-GAN, a custom GAN that learns pathological features from binary class labels alone, and LLIFT-DM, a diffusion-based inpainting pipeline conditioned on bounding-box masks via ControlNet.
Both approaches are evaluated on brain magnetic resonance imaging data derived from the Human Connectome Project. 
In evaluations, both achieve Fr\'echet Inception Distance scores, with respect to the real pathological distribution, that are comparable to the inter-class reference between healthy and pathological images in the given dataset.
Furthermore, qualitative inspection confirms the realism of lesion structures. 
The resulting benchmark datasets provide spatially controlled ground truth data for evaluating XAI methods in medical imaging.
\end{abstract}

\begin{IEEEkeywords}
Explainable Artificial Intelligence,
Generative Adversarial Networks,
Diffusion Models,
Magnetic Resonance Imaging,
Image Generation,
Benchmarking,
Ground-Truth Data
\end{IEEEkeywords}

\section{Introduction}

Machine Learning (ML) models increasingly inform clinical decisions in medical imaging. To deploy such models responsibly, it is often desired that their outputs are ``interpretable''.
Explainable AI (XAI) techniques have been proposed to address this desire \cite{vanderveldenExplainableArtificialIntelligence2022}, where the predominant paradigm is that of feature attribution.
However, due to the lack of formal XAI problem specifications, it is not clear what features are systematically highlighted by such methods in a given application context, and biases towards non-informative suppressor and salient image features have been reported \cite{wilmingTheory2023,clarkFeatureSalienceNot2026}.
Thus, it is necessary to validate XAI; however, such validation requires ground-truth data with precise locations of the discriminative features that contain actual information about the target to be predicted.
Using expert-annotated datasets can be a first step, but such an approach may miss lesions or contain incorrectly labeled healthy tissue, introducing unquantifiable biases into feature attribution evaluations \cite {wilming2022Scrutinizing,clarkXAITRISNonlinearImage2024,haufe2024position}. 
Existing synthetic benchmarks, on the other hand, rely on hand-crafted artificial anomalies, such as geometric shapes superimposed onto healthy images, producing perturbations that are insufficiently representative of real pathology~\cite{oliveira2024Bench}.
Both limitations compromise the reliability of assessments of feature attribution methods.

We propose Local Label-Informed Feature Transfer (LLIFT), a framework for generating semi-synthetic magnetic resonance imaging (MRI) brain images with controllable lesion placement that avoids these shortcomings.
The key idea is to learn the visual signature of pathology from weak supervision, driven by class membership or coarse spatial masks, rather than pixel-level segmentations or bounding boxes as labels.
We compare two instantiations of this framework: a Generative Adversarial Network (GAN) based approach (LLIFT-GAN) and a diffusion model-based approach (LLIFT-DM), highlighting their respective strengths and trade-offs.

\begin{figure}
    \centering
    \includegraphics[width=0.85\linewidth]{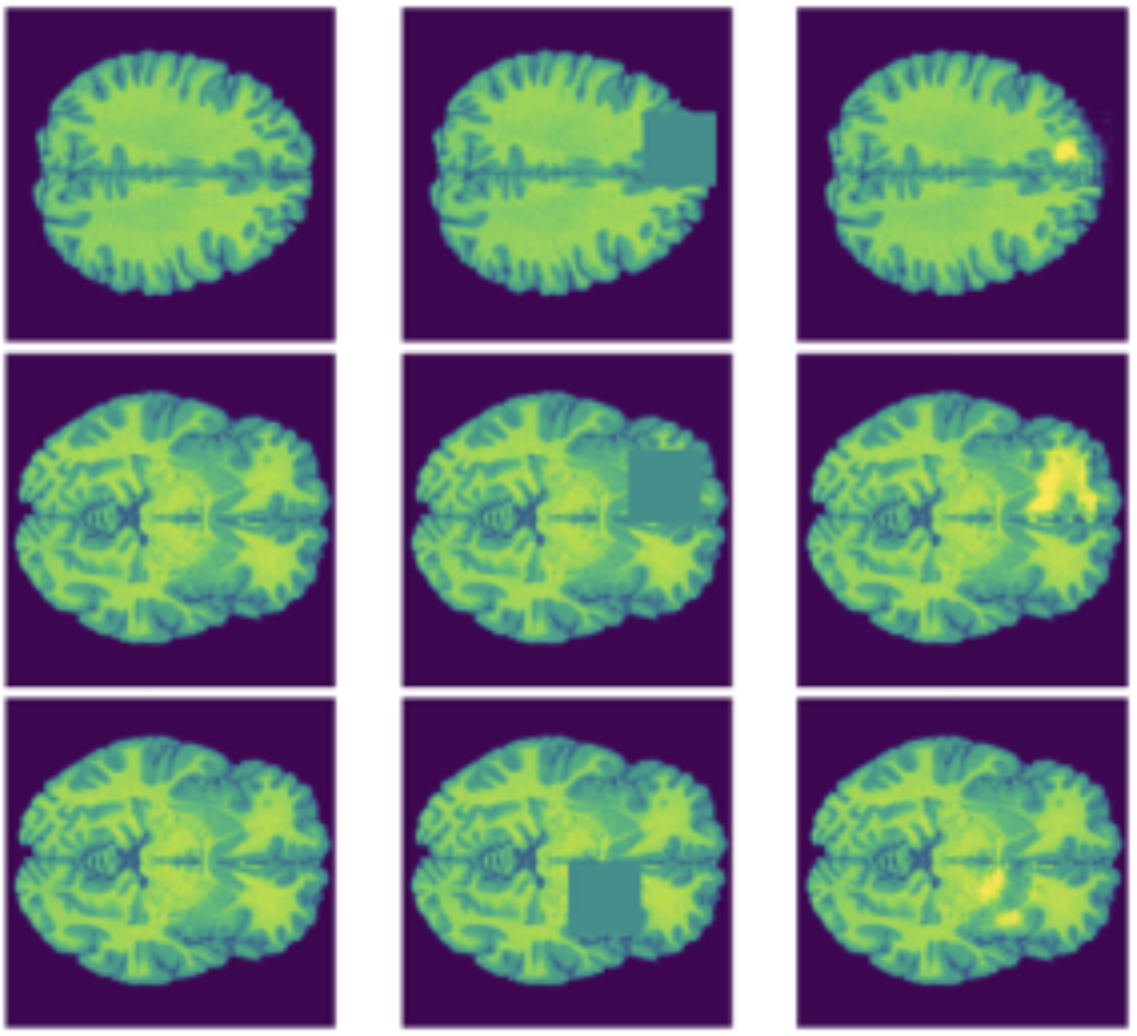}
    \caption{Examples of generated pathological images using LLIFT-GAN.
    From left to right column: Depicted are the original healthy images from the Human Connectome Project, in the middle column, healthy images with a specified mask signaling where lesions should be generated, and on the right, the resulting generated pathological images, where all pixels outside the mask area are identical to the healthy images.}
    \label{fig:gan-examples}
\end{figure}

\section{Related Work}

\paragraph*{Ground-truth-based XAI evaluation}
The lack of formal ground-truth feature attributions is a recognized problem for validating feature attribution methods, and several lines of work have proposed synthetic datasets to address this issue.
For image data, XAI-TRIS \cite{clarkXAITRISNonlinearImage2024} provides controllable synthetic scenarios that derive ground-truth feature importance from explicitly modeled data-generation processes.
In the medical domain, evaluation has commonly relied either on expert-annotated lesion segmentations, which are subject to labeling noise and inter-rater variability \cite{oliveira2024Bench,haufe2024position}, or on synthetic perturbations superimposed onto otherwise healthy slices \cite{oliveira2024Bench,galleeFunnyNodulesCustomizableMedical}.
LLIFT differs from these benchmarks in that it inserts clinically plausible lesion-like content directly into real MRI slices, while keeping the location of the discriminative feature known by construction, and does so without requiring pixel-level lesion segmentations during training.

\paragraph*{Lesion synthesis and inpainting in brain MRI}
Generative models for brain MRI have mainly fallen into two directions.
Pseudo-healthy synthesis aims at the inverse problem of removing lesions, where the BraTS Local Synthesis Inpainting Challenge \cite{kofler2023brain} establishes the standard benchmark, and methods such as FastSurfer-LIT \cite{pollak2025fastsurfer} and LRI-GAN \cite{liu2024lesion} build on diffusion models and GANs, respectively, to fill lesioned regions with healthy-looking tissue \cite{durrer2023denoising}.
Pathology synthesis, in contrast, transfers lesion-like content into healthy or otherwise lesion-free regions.
GAN-based approaches generate pathological volumes from full-anatomy segmentation maps \cite{shin2018medical} or from hand-crafted geometric priors, such as concentric circles, which are converted into tumor shapes by a learned network \cite{kim2021synthesis}.
More recent diffusion-based pipelines, such as multitask brain tumor inpainting \cite{rouzrokh2022multitask} and LeFusion \cite{zhang2025lefusion}, condition generation on bounding boxes or lesion-focused masks.
A common assumption across this body of work is that pixel-level lesion annotations are available during training, either as targets, conditioning inputs, or both.
LLIFT relaxes this assumption, where LLIFT-GAN uses only the binary healthy-vs-pathological class label, and LLIFT-DM uses bounding-box masks that are shuffled across images so that the conditioning signal is decoupled from the actual lesion location at training time.

\paragraph*{Conditional generation paradigms}
Conditional image generation in the context of XAI has been explored, for example, by \cite{goyal2019explaining}, which approximates specific causal concepts to generate causally related image features using a variational auto-encoder.
From an architectural perspective, LLIFT-GAN extends the DCGAN family \cite{radford2015unsupervised} with a dual-view discriminator and an output that is, by construction, limited to the conditioning mask, placing it within the broader literature on conditional image inpainting.
LLIFT-DM follows the recent paradigm of adapting large-scale pretrained latent diffusion models \cite{rombach2022high} to medical imaging through ControlNet-based fine-tuning \cite{zhang2023adding}.
RoentGen uses this approach for text-conditioned chest X-ray synthesis \cite{chambon2022roentgen}, and methods such as RePaint \cite{lugmayr2022repaint} use training-free diffusion inpainting.
In contrast to these works and \cite{goyal2019explaining} specifically, LLIFT-DM is, to our knowledge, the first to combine a foundation diffusion model with mask-conditioned fine-tuning specifically for the purpose of generating semi-synthetic pathological brain MRI images intended as XAI benchmark data, and to do so with an explicit post-hoc blending step that guarantees that pixels outside the conditioning mask remain identical to the healthy input.

\section{Dataset}

Both LLIFT variants are trained and evaluated on a derivative of the Human Connectome Project (HCP) \cite{vanessen2013}, which provides axial T1-weighted MRI slices of healthy young adults.
To obtain class-paired data, we use data from \cite{oliveira2024Bench}, in which smoothed-noise lesions with regular and irregular compactness are inserted into otherwise healthy slices, yielding aligned healthy/pathological pairs along with binary ground-truth masks.
The original 3D volumes are converted to 2D slices, cropped, and padded to a uniform $270\times 270$ resolution.
LLIFT-GAN operates at the native resolution, whereas for LLIFT-DM, we upscale the images to $512\times 512$ to match the input format of the pretrained Stable Diffusion model.
We emphasize that the ground-truth lesion masks are used only for evaluation and for constructing the conditioning masks at inference; neither model receives the true pixel-level lesion location during training.

\section{Methodology}

Two guiding principles steer LLFIT's design: a lesion's location must be specified by the user, and its visual characteristics must be learned from data.
% The two LLIFT variants are designed around the guiding principle that the location of a lesion in the output must be controlled by the user, while the visual appearance of that lesion must be learned from data rather than specified by hand.
Both variants therefore share the same prediction interface, in which a 2D slice of a healthy brain MRI is combined with a spatial mask specifying where lesion-like content should appear, and the generator fills only the masked region.
The two paradigms differ in how they realize the generation process.
LLIFT-GAN directly trains a discriminator on the binary healthy-vs-pathological label and lets the adversarial training procedure transfer class-informative features into the masked region.
LLIFT-DM is an indirect approach, where, via spatial conditioning, a large-scale pretrained diffusion model is fine-tuned with mask-aware inpainting, trading direct discriminative supervision for stronger natural-image priors.

\subsection{LLIFT-GAN}

LLIFT-GAN employs a GAN architecture to learn pathological features from binary class labels; no bounding boxes or other spatial annotations are provided during training. 
The discriminator implicitly discovers those features that distinguish healthy from pathological images and, through adversarial training, teaches the generator to reproduce them within externally specified mask regions. 
The architecture extends DCGAN \cite{radford2015unsupervised} with an encoder-decoder generator whose output is confined to the masked area.

\paragraph*{Generator}
The generator $G_\theta$ takes as input a healthy image $x^h\in\mathbb{R}^{W\times H}$, a binary mask $m\in\{0,1\}^{W \times H}$ defining a quadratic bounding box at a randomized location, and a noise vector $z$.
The image branch encodes $x^h$ through four convolutional layers with ReLU activations and batch normalization; a parallel branch upsamples $z$ to the spatial resolution; the two branches are concatenated and decoded by transposed convolutions.
The decoder output is restricted to the mask region by construction, so that the generator's mapping reduces to
\begin{equation}
G_\theta(x^h,m,z) = (1-m) \odot x^h + m \odot \tilde{G}_{\theta}(x^h, m, z),
\label{eq:gan-G}
\end{equation}
where $\tilde{G}_\theta$ is the raw decoder output.
This architectural choice prevents any post-hoc blending.

\paragraph*{Discriminator}
A discriminator is used to determine whether a pathological image is real or generated.
% Within the GAN paradigm, a discriminator is used to determine whether a generated image is real or fake; in our context, this translates to discriminating between real and generated pathological images.
We found that a straightforward single-scale discriminator was insufficient.
As a global critic, it captures whether the overall image resembles a real pathological scan and is largely insensitive to small lesion-sized edits, whereas a local critic, restricted to the bounding box, loses the anatomical context that grounds the plausibility of a given patch.
We therefore use a dual-view discriminator that processes the image at two views and sums the resulting features before the final scalar projection
\begin{equation}
D_\phi(x,m) = s \left(W \big[h_{\phi}(x) + h_{\phi}(x \odot(1-m)) \big] \right),
\end{equation}
where $h_\phi$ is a single convolutional feature extractor (the same weights for both views) followed by a minibatch-discrimination layer \cite{salimans2016improved}, $s$ is the sigmoid, and $W$ is a linear projection to a scalar.
The first term captures the full image, and the second term masks out the bounding box, so the critic is also forced to read evidence from the surrounding tissue.
For a real pathological image, the lesion is somewhere in the slice, and because the conditioning mask $m$ is sampled independently from the lesion location, the lesion typically survives the $x \odot (1-m)$ operation, and still carries pathological information.
For a generated image $G_\theta(x^h,m,z)$, the generator's contribution lives entirely inside $m$, so multiplying by $(1-m)$ erases it and the masked-out view collapses to a clean, healthy image.
The second term, therefore, acts as a generator-independent prior that pushes $D_\phi$ to label samples ``pathological'' whenever the unmasked context appears pathological.
The generator can only adapt through the full-image view, which forces it to produce lesion content compelling enough at the global scale to override that prior.

\paragraph*{Optimization problem}
With $x^u \sim P_u$ drawn from real pathological images, $x^h \sim P_h$ from real healthy images, $z \sim N(0,I)$, and $m$ sampled from the distribution of randomly placed quadratic masks, LLIFT-GAN solves the smoothed adversarial objective: $\min_{\theta}\max_{\phi}
(1-\alpha) \, \mathbb{E}_{x^u, m} [ \varphi ] + \mathbb{E}_{x^h, z, m}[ \psi ]$,
%
% \begin{align}
% \min_{\theta}\max_{\phi} \; &
% (1-\alpha) \, \mathbb{E}_{x^u, m} [ \varphi ] + \mathbb{E}_{x^h, z, m}[ \psi ],
% \label{eq:gan-obj}
% \end{align}
%
where $\varphi = \log D_\phi(x^u,m)$, $\psi = \log(1 - D_\phi(G_\theta(x^h, m, z), m))$, and $\alpha\in(0,1)$ implements one-sided label smoothing on the real image branch.

\paragraph*{Training}
Training alternates between updating $D_\phi$ and $G_\theta$ as in the original GAN formulation \cite{goodfellow2014generative}, optimized with Adam ($\beta_1$ ramped from $0.9$ to $0.5$) and a learning rate decayed from $10^{-3}$ to $10^{-5}$.

\subsection*{LLIFT-DM}

GANs deliver direct class supervision but must learn the full image distribution from a few thousand medical scans, which, in our experiments, frequently led to instability, mode collapse, and a general sensitivity to architectural details.
Large-scale pretrained latent diffusion models \cite{rombach2022high} avoid this bottleneck by transferring a strong natural-image prior, leaving the fine-tuning task to specialize the model toward brain MRIs and lesion-shaped local edits rather than relearning low-level image statistics.
LLIFT-DM leverages the generative capacity of diffusion models by finetuning Stable Diffusion \cite{rombach2022high} with a ControlNet \cite{zhang2023adding} inpainting module.
Unlike LLIFT-GAN, this approach conditions on bounding-box masks rather than purely binary labels. During training, lesion masks from the training dataset are shuffled across samples to decouple lesion location from image content, and only the UNet and ControlNet components are finetuned while all other modules remain frozen.
A post-hoc blending step replaces only the masked pixels in the original healthy image with the generated content, guaranteeing that unmasked regions remain identical to the input.

\paragraph*{Conditioning}
For each training sample, the ground-truth lesion mask is first shuffled to a different image in the batch, then its smallest enclosing rectangle $B=[x_{\min},x_{\max},y_{\min},y_{\max}]$ is computed, expanded by a fixed padding $r$ to $B_r$, and turned into the binary mask $m \in \{0,1\}^{W\times H}$ that is $1$ inside $B_r$ and $0$ elsewhere.
The ControlNet conditioning image $c \in \mathbb{R}^{W \times H} $ is given by $c_{ij}= -1$ if $(i,j) \in B_r$ otherwise $c_{ij} = x_{ij}$,
%
% \begin{equation}
% c_{ij} = \begin{cases} 
% -1.0 & (i,j) \in B_r \\
% x_{i j} & \text{otherwise,}
% \end{cases}
% \end{equation}
%
i.e., the masked region is set to a constant placeholder value while the surrounding context is preserved.
A fixed text prompt ``\textit{A brain MRI with a lesion}'' is tokenized and encoded by the frozen CLIP text encoder \cite{rombach2022high} to yield a text-embedding $p$ shared across all samples.
The embedding $p$ serves primarily as a placeholder satisfying the architecture's textual-conditioning interface, with the spatial mask acting as the dominant conditioning signal.

\paragraph*{Optimization problem}

Let $\mathcal{E}$ denote the frozen VAE encoder, so that $z_0=\mathcal{E}(x^u)$ is the latent representation of a pathological training image.
A timestep $t \sim \mathcal{U} \{1, \dots, T\}$ is drawn, Gaussian noise $\epsilon \sim N(0, I)$ is sampled, and the noised latent follows the flow-matching schedule $z_t = (1- \sigma_t) z_0 + \sigma_t \epsilon.$
%
% \begin{equation}
% z_t = (1- \sigma_t) z_0 + \sigma_t \epsilon.
% \end{equation}
%
Writing $\epsilon_{\theta,\phi}$ for the noise prediction produced by the UNet (parameters $\theta$) together with the ControlNet residuals (parameters $\phi$), and recalling that the conditioning $(p,c)$ is constructed from the shuffled-mask procedure above, LLIFT-DM solves
\begin{equation}
\min_{\theta,\phi} \; \mathbb{E}_{x^u,\,t ,\,\epsilon, \,m} \Big[\big \| \epsilon -\epsilon_{\theta,\phi} \big(z_t, t, p, c \big) \big\|_2^{2} \Big],
\label{eq:dm-obj}
\end{equation}
where the expectation over $m$ encodes the fact that the conditioning mask is sampled independently of $x^u$.
This decoupling forces $\epsilon_{\theta,\phi}$ to learn the visual signature of pathology rather than memorizing where lesions occur in the training set.

\paragraph*{Training}

During training, the VAE, CLIP text encoder, and scheduler remain frozen, and only the UNet ($\theta$) and ControlNet ($\phi$) branches are updated.
Training runs for 8 epochs at learning rate $10^{-5}$ with cosine annealing (3 cycles) and a linear warm-up over the first $10\%$ of steps.

\paragraph*{Inference and blending}

At inference time, between one and four bounding boxes are sampled within the brain mask of a healthy test image, and the diffusion pipeline generates content for the masked region only.
Because the unmasked stochastic latent decoding can still introduce sub-pixel changes outside $B_r$, we apply an explicit post-hoc blending mechanism
\begin{equation}
\hat{x}^u_\text{blended}=(1 - m) \odot x^h + m \odot \hat{x}^u,
\end{equation}
where $\hat{x}^u$ is the generated pathological image.
This blending mechanism ensures that pixels outside the mask are identical to the healthy input.

Both approaches guarantee that healthy regions remain untouched.
LLIFT-GAN restricts generator output to the mask via architectural constraints, while LLIFT-DM enforces this through explicit post-hoc blending.

\subsection{Evaluation Metric}
\label{sec:metric}

We evaluate both paradigms with a single quantitative metric, the Fr\'echet Inception Distance (FID) \cite{heusel2017gans}, complemented by qualitative inspection.
FID models the distributions of Inception-v3 features from two image sets as multivariate Gaussians and reports the squared Wasserstein-2 distance between them.
The meaning of the raw FID values depends on the dataset's natural inter-batch variance.
We therefore report every FID alongside two reference values: an \emph{intra-class} baseline, computed between two distinct batches drawn from the same class, and an \emph{inter-class} baseline, computed between a healthy batch and a real pathological batch.
A generator that reaches the inter-class reference has, by definition, produced a batch that is distributionally as different from healthy data as real pathological data is.

% ---------------------------------------------------------------
\section{Results}

We evaluate each paradigm from two angles, as motivated in Section~\ref{sec:metric}: qualitative inspection of generated samples, which assesses whether lesions are visually plausible, and FID against the natural inter-class reference, which assesses whether the generated samples are distributionally close to real pathological scans rather than healthy ones.

\subsection{Qualitative Evaluation}

Figure~\ref{fig:gan-examples} shows representative LLIFT-GAN outputs.
The generated lesions are confined to the bounding boxes, exhibit organic shape variation, and blend smoothly with the surrounding.
The model has clearly learned that a lesion is a brighter, irregularly shaped region rather than a geometric blob.
In some cases, artifacts can appear with the given bounding box, causing the generated lesion to be less pronounced (see Figure \ref{fig:gan-negative-examples} in the Appendix).

LLIFT-DM produces a comparable mix of successful and less successful generations (See Appendix \ref{app:dm-examples} Figure \ref{fig:dm-examples}).
In successful cases, lesions integrate seamlessly with the background texture, e.g., the first two rows of Figure \ref{fig:dm-examples}.
In less successful cases, the last row shows lesions characterized by the absence of matter or other non-lesion-like brain structures.

\subsection{Quantitative Similarity Metrics}

In Table~\ref{tab:fid}, we report all FID values. 
LLIFT-GAN was evaluated only against its inter-class reference because its single output mode (architecturally mask-confined generation) does not afford a meaningful blended-vs-full split, while LLIFT-DM was also evaluated on the two distinct output modes (raw and blended).

\begin{table}[h!]
\centering
\caption{FID values for both LLIFT paradigms. The upper block reports reference values computed between real-data batches; the lower block reports values involving generated outputs. Dashes mark settings not evaluated for the given paradigm.}
\label{tab:fid}
\begin{tabular}{lcc}
\toprule
Comparison setting & LLIFT-GAN & LLIFT-DM \\
\midrule
\multicolumn{3}{l}{\textbf{Reference values} (real data only)} \\
Healthy vs.\ healthy (intra-class)      & $27.87$ & --- \\
Healthy vs.\ pathological (inter-class) & $41.75$ & $5.84$ \\
\midrule
\multicolumn{3}{l}{\textbf{Generated outputs}} \\
% Healthy vs.\ blended generated          & ---     & $4.78$ \\
% Healthy vs.\ fully generated            & ---     & $8.20$ \\
Pathological vs.\ blended generated     & $41.69$ & $7.61$ \\
% Pathological vs.\ fully generated       & ---     & $9.74$ \\
Healthy vs.\ blended generated          & ---     & $4.78$ \\
Healthy vs.\ fully generated            & ---     & $8.20$ \\
% % Pathological vs.\ blended generated     & $41.69$ & $7.61$ \\
Pathological vs.\ fully generated       & ---     & $9.74$ \\
\bottomrule
\end{tabular}
\end{table}
%
%
% \begin{table}[h!]
% \centering
% \caption{FID values for LLIFT-DM.}
% \label{tab:fid-appendix}
% \begin{tabular}{lcc}
% \toprule
% Comparison setting & LLIFT-GAN & LLIFT-DM \\
% \midrule
% \multicolumn{3}{l}{\textbf{Generated outputs}} \\
% Healthy vs.\ blended generated          & ---     & $4.78$ \\
% Healthy vs.\ fully generated            & ---     & $8.20$ \\
% % Pathological vs.\ blended generated     & $41.69$ & $7.61$ \\
% Pathological vs.\ fully generated       & ---     & $9.74$ \\
% \bottomrule
% \end{tabular}
% \end{table}
%

\paragraph*{LLIFT-GAN}
The LLIFT-GAN reaches an average FID of $41.69$ between generated and real pathological batches.
This value is comparable to the inter-class reference of $41.75$, meaning the generated batch is distributionally as far from real pathological data as real healthy data is.
A direct interpretation would be that this is no improvement over the healthy baseline.
On the other hand, the GAN modifies only a small bounding-box region of an otherwise healthy image, so the overall image-level FID is necessarily dominated by the unchanged healthy background.
Visual inspection (Figure~\ref{fig:gan-examples}) suggests that the lesion patches themselves are clearly pathological and structurally distinct from healthy tissue.
% For LLIFT-DM, we provide additional comparison settings for the generated outputs with corresponding FID scores in Table \ref{tab:fid-appendix} in the appendix. 

\paragraph*{LLIFT-DM}
The blended generated images are close to the healthy inputs (FID $4.78$), which is the expected effect of the blending step.
Outside the lesion bounding box, the output is identical to the input, so the only source of distributional disagreement is the small generated patch.
Compared with the real pathological distribution, the blended outputs achieve an FID of $7.61$, slightly above the inter-class reference of $5.84$.
This suggests the diffusion-generated batch is marginally further from real pathological data than real healthy data is, with the same small-mask caveat as for LLIFT-GAN.
Fully generated images, in which the entire frame is re-synthesized, exhibit systematically larger distributional distance, confirming that the blending step is an essential part of the pipeline.

\section{Discussion}

When feature attribution methods are used to interpret machine-learning model decisions, the underlying assumption is that strongly attributed features are informative for the prediction task, that the image regions a method highlights contain content whose distribution actually differs between classes \cite{wilmingTheory2023, wilming2022Scrutinizing, haufe2024position}.
Under that assumption, attribution maps could in principle support downstream goals such as model and data diagnostics and scientific discovery \cite{haufe2024position}.
The assumption is widely held, yet it remains empirically unverified in most realistic settings, because benchmarks with known informative features are scarce. 
With the LLIFT framework, we propose a design that precisely provides such benchmark data to test the informativeness assumption directly.
And the benchmark design specifically represents a realistic scenario, where ``realistic'' refers to the structural fidelity of the data setting with real anatomical backgrounds and an anomaly-classification-style prediction task, compared to clinical-pathological realism, and we note that expert reader studies would become essential were this approach extended toward clinical data augmentation or deployment

With LLIFT-GAN, we propose an instantiation of LLIFT that learns class-discriminative features via adversarial learning on a binary label, with an explicit training signal for the discriminability objective.
LLIFT-DM, on the other hand, inherits the data efficiency and visual fidelity of large-scale pretrained diffusion models. 
During fine-tuning, it conditions on a bounding box rather than a binary class label, so its supervision is spatial rather than discriminative.

Both LLIFT-GAN and LLIFT-DM successfully transfer pathological content into a user-specified region without ever seeing pixel-level lesion annotations, and both reach distributional similarity to the real pathological class that is on the order of the natural inter-class reference (Table~\ref{tab:fid}).
The benchmark datasets that can result, consisting of a healthy input image, a binary mask, and a synthetic pathological version that is identical to the input outside the mask, have the structure required for ground-truth evaluation for feature attribution methods, where the discriminative region is, by construction, known.
For evaluating attribution methods, the relevant property is that the spatial location of the informative feature is known, and both LLIFT variants satisfy this property by construction.

Several limitations remain.
LLIFT-GAN training is rather unstable and depends on random seeds.
Its dual-view discriminator applies the $(1-m)$ masking symmetrically to real and generated inputs, so the random conditioning mask occasionally overlaps with the actual lesion in a real sample and partially erases it.
However, only masking the generated image might be disadvantageous.
Following \cite{karras2020training, zhao2020OnLeveraging}, masking only the generated images would cause the discriminator to ultimately distinguish masked vs. non-masked images, degrading generation performance. 
LLIFT-DM trains only on pathological images, and including healthy images with prompt-level distinction (``healthy brain MRI'' vs.\ ``MRI with a lesion'') could sharpen the learned lesion signature.
Also, for both models, replacing the synthetic lesions with real clinical pathology is the natural next step.
The image-level FID is limited as an evaluation metric when only a small region of the image is modified, regardless of how realistic that region appears.
Metrics operating at the scale of the edited region could improve quantification of generation quality.
Patch-level or mask-restricted variants of FID, or class-conditional metrics computed on a separately trained pathology classifier, would be more sensitive to the actual content of the lesion patch and are a natural direction for future evaluation protocols.

\section{Conclusion}

We presented LLIFT, a framework for generating semi-synthetic brain MRIs with spatially controlled lesion placement under weak supervision, and instantiated it with both a GAN (LLIFT-GAN) and a diffusion model (LLIFT-DM).
Both reach Fr\'echet Inception Distance scores comparable to the natural inter-class reference between real healthy and pathological images, and both restrict their edits to user-specified masks.
They differ in the nature of their supervision: LLIFT-GAN uses the adversarial class signal, while LLIFT-DM uses spatial mask conditioning.
The images generated by both LLIFT variants provide ground truth data for evaluating whether attribution methods correctly identify pathology-relevant regions, since lesion locations are precisely defined by the input masks.
The framework addresses data scarcity for pathological images and enables standardized, reproducible benchmarking of feature attribution techniques.

\section{Acknowledgment}
This work is supported by Project 22HLT05 MAIBAI, which has received funding from the European Partnership on Metrology, co-financed from the European Union’s Horizon Europe Research and Innovation Programme and by the Participating States.

% ---------------------------------------------------------------
\bibliographystyle{ieeetr}
\bibliography{references}

\appendix

\section{LLIFT-GAN Generation Examples}

\begin{figure}
    \centering
    \includegraphics[width=0.85\linewidth]{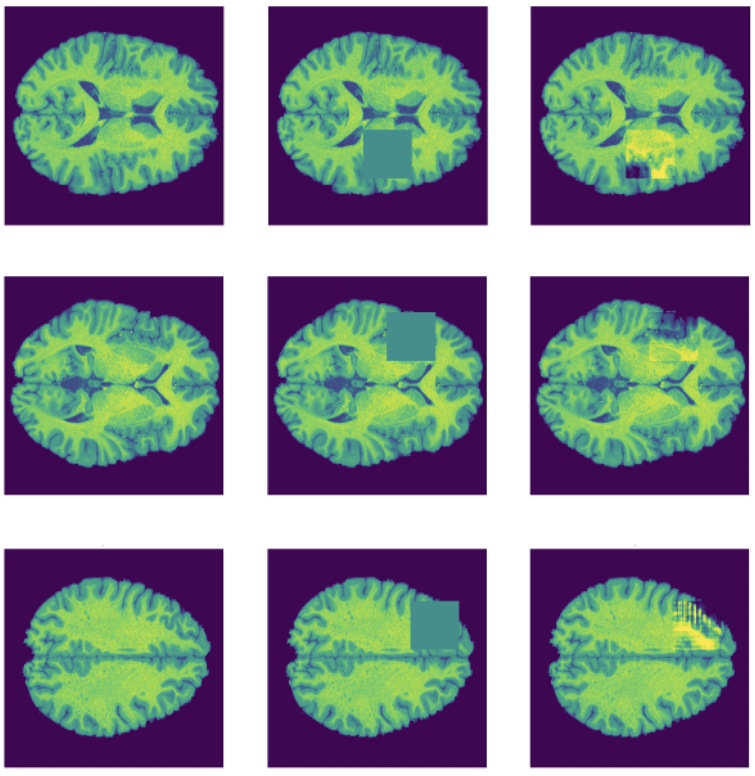}
    \caption{Examples of generated pathological images with artifacts using LLIFT-GAN.
    From left to right column: Depicted are the original healthy images from the Human Connectome Project, in the middle column, healthy images with a specified mask signaling where lesions should be generated, and on the right, the resulting generated pathological images, where all pixels outside the mask area are identical to the healthy images.}
    \label{fig:gan-negative-examples}
\end{figure}

\section{LLIFT-DM Generation Examples}
\label{app:dm-examples}

\begin{figure}[h!]
    \centering
   \includegraphics[width=0.85\linewidth]{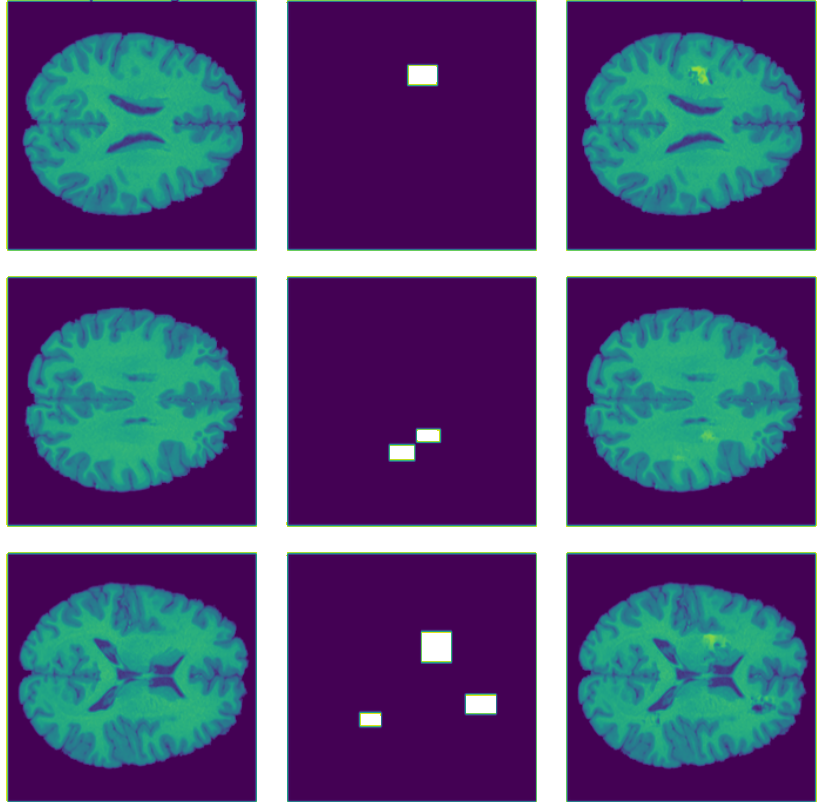}
    \caption{Examples of generated pathological images using LLIFT-DM.
    From left to right column: original healthy images from the Human Connectome Project; bounding-box masks specifying where lesions should be inserted; the resulting blended pathological images, where all pixels outside the mask area are identical to the healthy inputs by construction.}
    \label{fig:dm-examples}
\end{figure}

% \section{Additional Quantitative Performance Results}

% \begin{table}[h!]
% \centering
% \caption{FID values for both LLIFT paradigms. Reported are values involving generated outputs. Dashes mark settings that were not evaluated for the given paradigm.}
% \label{tab:fid-appendix}
% \begin{tabular}{lcc}
% \toprule
% Comparison setting & LLIFT-GAN & LLIFT-DM \\
% \midrule
% \multicolumn{3}{l}{\textbf{Generated outputs}} \\
% Healthy vs.\ blended generated          & ---     & $4.78$ \\
% Healthy vs.\ fully generated            & ---     & $8.20$ \\
% Pathological vs.\ blended generated     & $41.69$ & $7.61$ \\
% Pathological vs.\ fully generated       & ---     & $9.74$ \\
% \bottomrule
% \end{tabular}
% \end{table}

\end{document}